\documentclass[journal]{IEEEtran}

\usepackage{cite}
\usepackage{amsmath,amssymb,amsfonts}
\usepackage{tabularx}
\usepackage{multirow}
\usepackage{booktabs}
\usepackage{booktabs} 
\usepackage{multirow} 
\usepackage{algorithmic}
\usepackage{graphicx}
\usepackage{subcaption} 
\usepackage{textcomp}
\usepackage{xcolor}
\usepackage{url} 
\usepackage{array}
\usepackage{tabularx}
\usepackage[ruled,vlined,linesnumbered]{algorithm2e}
\def\BibTeX{{\rm B\kern-.05em{\sc i\kern-.025em b}\kern-.08em
    T\kern-.1667em\lower.7ex\hbox{E}\kern-.125emX}}
\begin{document}

\title{Physics-Aware Random Walk Fingerprints for Scalable Power Grid Graph Classification }

\author{Adnan Anwar,~\IEEEmembership{Member,~IEEE}
\thanks{Associate Professor A. Anwar is with the School of Information Technology and a member of the Deakin Cyber Research \& Innovation Hub, Deakin University, Melbourne, Australia. (see https://experts.deakin.edu.au/50139-adnan-anwar), Email: adnan.anwar@deakin.edu.au}
\thanks{Manuscript received September 7, 2026; revised MMM DD, YYYY.}}

\maketitle

\begin{abstract}
Cyber-physical infrastructures like power systems generate attributed graph data in which
system-level outcomes depend on both network connectivity and operational
interaction patterns. Recent benchmarks such as \textit{PowerGraph}
provide large collections of power-grid graphs for cascading-failure
classification.
Graph neural networks (GNNs) achieve strong predictive performance on this
task, but typically require end-to-end training and model-specific tuning, while
their latent representations can be difficult to relate to physically meaningful
propagation patterns. Random Walk Fingerprints (RWF) offer a scalable and
interpretable alternative, but existing variants primarily emphasise topology
and node-level information, leaving grid-relevant operational edge states in the walk dynamics.

We propose Multi-Channel Physics-Aware Random Walk Fingerprints
(MC-PA-RWF) for power systems, a lightweight graph-level representation framework that introduces
physical edge states into random-walk propagation. The method constructs
multiple edge-weighted channels from domain-relevant attributes, extracts a
channel-specific fingerprint from each weighted graph, and concatenates the
resulting vectors into a compact representation. Experiments on three
\textit{PowerGraph} benchmark systems show substantial improvements over
topology-only RWF and competitive balanced accuracy against strong GNN
baselines, including Graph Convolutional Networks (GCN), Graph Attention
Networks (GAT), Graph Isomorphism Networks with edge features (GINE), and
Transformer-based Graph Convolutional Networks (TransformerConv). At the
largest evaluated settings, the node-edge extension MC-PA-RWF+ achieves around
\textbf{98.04\%} $-$ \textbf{{99.32\%}} balanced accuracy and improves failure-class F1 over the
strongest GNN baseline by \textbf{1.60} $-$ \textbf{5.84} percentage points, with statistically
significant gains across all three systems. These results show MC-PA-RWF
as a scalable and interpretable approach for cyber-physical graph mining for power systems.
\end{abstract}


\section{Introduction}
\label{sec:introduction}

Modern power systems generate large volumes of networked operational data from
monitoring, protection, and control. In a power system, nodes represent buses,
generators, loads, or substations, while edges represent transmission lines or
transformers with operating states (e.g., flows, limits, and overload status)~\cite{anwar2022measurement}.
System security is therefore not determined by topology alone: it also depends
on how stress, capacity limitations, and failure effects propagate through
attributed electrical connections. Each operating condition can be modelled as
an attributed power-grid graph and the target label can reflect system-level
security or stability~\cite{powergraph}. Such applications require
graph-classification methods that are scalable, interpretable, and sensitive to
power-system-relevant node and edge attributes~\cite{gorka_pscc24}.

Recent benchmarks have enabled systematic evaluation of graph-learning methods
for power grid analytics. In particular, \textit{PowerGraph} provides
power-system graph datasets across multiple tasks and grid sizes
\cite{powergraph}. Each operating scenario is represented using realistic
topology, node features, edge features, and task-level labels, supporting
problems such as power-flow prediction, optimal operation, and cascading-failure
analysis. Unlike many conventional graph-classification benchmarks from
chemistry, biology, and social networks, PowerGraph highlights settings in which
transmission-line operating states and other physically grounded edge
attributes can directly influence system-level outcomes for power systems.

Graph neural networks (GNNs) provide a natural modelling framework for these
datasets. Architectures such as Graph Convolutional Networks (GCN), Graph
Attention Networks (GAT), Graph Isomorphism Networks with edge features (GINE),
and TransformerConv models learn graph representations by propagating
information across connected components
\cite{gcn,gat,gine,transformerconv}. However, end-to-end neural training requires
model-specific optimisation and hyperparameter tuning, while the resulting
latent representations can be difficult to relate directly to physical
interaction patterns. This motivates complementary graph-mining approaches that
remain lightweight, reproducible, and interpretable.

Random Walk Fingerprints (RWF) provide a scalable non-neural alternative
\cite{rwf}. RWF transforms each graph independently into a fixed-length vector
by summarising multi-step random-walk connectivity within and between
structurally aligned node groups. The resulting fingerprint is interpretable,
compatible with conventional classifiers such as support vector machines, and
scales linearly with the number of graphs. However, the standard RWF transition
process is primarily topology driven. Although node attributes can be included
through additional feature interactions, edge attributes do not directly shape
the walk dynamics.

This limitation is important for cyber-physical graphs like power system or power grid topological graphs. Two transmission lines
with the same topological role may have different operational meanings because
of their loading levels, residual capacities, coupling strengths, or overload
conditions. Treating them as identical links in a binary adjacency matrix can
remove information that is central to failure propagation. To address this gap,
we propose {Multi-Channel Physics-Aware Random Walk Fingerprints
(MC-PA-RWF)}. The method represents each graph using multiple edge-weighted
channels defined over the same topology. Each channel captures a physically
meaningful interaction pattern, such as flow intensity, capacity margin,
coupling strength, or overload severity of electric lines. Channel-specific RWF vectors are then
extracted and concatenated into a compact graph-level representation. This extends RWF from a topology-driven graph descriptor to an edge-attributed representation that captures propagation behaviour induced by physics-aware edge states in cyber-physical power system systems.

In this paper, we evaluate MC-PA-RWF on the \textit{PowerGraph} graph-level
cascading-failure classification task. In modern power systems, operating conditions
change dynamically due to renewable generation, demand variability, feeder reconfiguration, and contingency events. Therefore, graph-level risk assessment requires
models that can capture not only network topology, but also the physical state of
transmission lines. The proposed method is compared with topology-only and
node-feature RWF benchmark models, channel-specific PA-RWF variants, and benchmark neural graph
models including GCN, GAT, GINE, and TransformerConv. 
The results show that physics-aware edge channels improve over topology-only RWF and remain competitive with strong GNN baselines while preserving interpretable power grid graph structure.

The main contributions of this paper are as follows:
\begin{itemize}

\item \textbf{Physics-aware random-walk fingerprints:}
We extend Random Walk Fingerprints~\cite{rwf} beyond topology and node attributes
by incorporating transmission-line operating states into the random-walk
transition dynamics. This enables graph-level representations to reflect
power grid operating conditions, where stability and cascading-failure risk are
influenced by how electrical stress propagates through physically connected
network components.

\item \textbf{Multi-channel modelling of transmission-line conditions:}
We propose {MC-PA-RWF}, a lightweight framework that represents each
power grid operating point through multiple edge-weighted channels over the same
network topology. Each channel encodes a power-system relationship, such as flow
intensity, capacity margin, coupling strength, or overload severity. This allows
random walks to follow electrically meaningful interaction strengths while
preserving the scalability and interpretability of RWF.

\item \textbf{Validation on power grid cascading-failure benchmarks:}
Experiments on IEEE24, IEEE39, and UK benchmark systems from \textit{PowerGraph}
show that the proposed method improves substantially over topology-only RWF,
remains competitive with strong GNN baselines~\cite{gcn,gat,gine,transformerconv},
and improves failure-class detection. The channel analysis further reveals which
transmission-line operating state measurements contribute most strongly to cascading-failure
classification, providing interpretable insight into power grid vulnerability
patterns.
\end{itemize}

\begin{figure}
\centering
\includegraphics[width=0.48\textwidth]{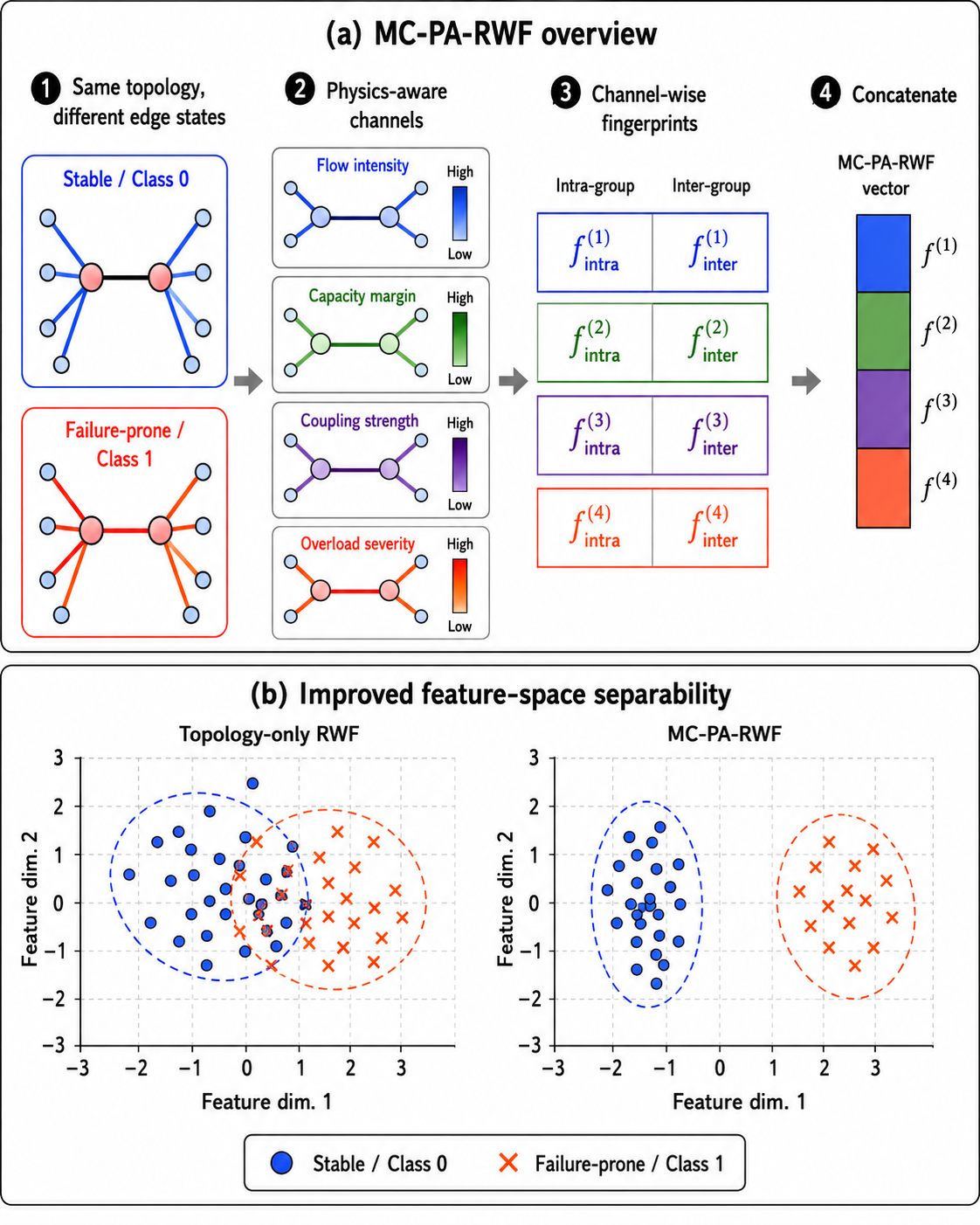}
\caption{Overview of the proposed MC-PA-RWF framework.}
\label{fig:overview}
\vspace{-0.8em}
\end{figure}

\textbf{Conceptual Overview:}
MC-PA-RWF is motivated by the observation that power-grid operating conditions
and cascading-outage risk can depend on transmission-line states and
edge-state-driven propagation patterns rather than topology alone.
As illustrated in Fig.~\ref{fig:overview}~(a), two operating scenarios may share the
same graph structure but differ in line conditions, leading to different
system-level outcomes. MC-PA-RWF captures this effect by constructing multiple
physics-aware edge-weighted channels, extracting channel-wise random-walk
fingerprints, and concatenating them into a unified graph representation.
The channels encode complementary physics-driven interaction patterns, including flow
intensity, capacity margin, coupling strength, and overload severity, obtained from the edge weights.
Fig.~\ref{fig:overview}~(b) schematically illustrates how the
multi-channel representation can improve feature-space separability over topology-only RWF~\cite{rwf}.

\textbf{Organisation of the Paper:}
Section~\ref{sec:related_work} reviews related work, followed by the RWF
preliminaries in Section~\ref{sec:preliminaries}. Section~\ref{sec:method}
presents the proposed physics-aware multi-channel random-walk fingerprint
framework. Section~\ref{sec:experimental_setup} describes the datasets, baseline
methods, and evaluation metrics, followed by the experimental results and
analysis in Section~\ref{sec:results}. Finally, Section~\ref{sec:conclusion}
concludes the paper.

\section{Related Work}
\label{sec:related_work}

\subsection{Graph Classification Methods}

Graph classification aims to assign a label to an entire graph by capturing 
discriminative structural and attribute patterns. Classical approaches include 
graph kernels, such as the Weisfeiler--Lehman kernel, neighbourhood hash kernel, 
GraphHopper kernel, and random-walk kernel~\cite{wl_kernel,nh_kernel,graphhopper,
random_walk_kernel}. These methods compare graphs using subtree patterns, local 
neighbourhood structures, shortest paths, or walk-based similarities. Although 
effective on conventional graph benchmarks, many kernel methods require pairwise 
graph comparisons, leading to quadratic scaling with the number of graphs. 
Fixed-length graph descriptors, including NetLSD, FGSD, and NetSimile, provide a 
more scalable alternative by summarising each graph independently using spectral, 
distance-based, or statistical features~\cite{netlsd,fgsd,netsimile}. However, 
generic descriptors may not fully preserve the domain-specific interaction 
patterns that determine system-level outcomes in cyber-physical networks.

\subsection{Random-Walk-Based Graph Representations}

Random walks provide a natural mechanism for characterising graph connectivity 
because they capture how information, influence, or failure effects may propagate 
through a network. Traditional random-walk kernels compare pairs of graphs by 
counting matching walks, but their computational cost can become prohibitive for 
large graph collections~\cite{random_walk_kernel}. Random Walk Fingerprints (RWF) 
address this limitation by transforming each graph independently into a compact 
fixed-length vector~\cite{rwf}. RWF partitions nodes into structurally aligned 
groups and summarises multi-step random-walk connectivity within and between these 
groups. The resulting representation is invariant to node ordering, interpretable 
at the level of node groups and walk lengths, and compatible with conventional 
classifiers such as support vector machines. Unlike pairwise graph kernels, RWF 
scales linearly with the number of graphs. Its original formulation primarily 
uses topology-based symmetrically normalised walk (propagation) matrices, with an optional mechanism for 
incorporating node-feature interactions.

\subsection{Graph Learning for Cyber-Physical Energy Systems}

Graph neural networks (GNNs) have become an important modelling tool for 
cyber-physical energy systems because power grids can be naturally represented 
as graphs 
whose nodes denote grid components, such as buses, generators, and loads, and
whose edges denote physical connections, such as transmission lines. Message-passing architectures, including Graph Convolutional 
Networks (GCN), Graph Attention Networks (GAT), Graph Isomorphism Networks with 
edge features (GINE), and TransformerConv models, learn task-specific 
representations by aggregating information across connected components 
\cite{gcn,gat,gine,transformerconv}. {\textit{PowerGraph}}~\cite{powergraph} provides a benchmark for 
evaluating such models on attributed power-grid data, including power-flow, 
optimal-power-flow, and cascading-failure analysis tasks. These 
benchmarks demonstrate the effectiveness of neural graph models for power-system 
analytics. However, end-to-end GNN training generally requires model-specific 
optimisation and hyperparameter tuning, while the resulting latent 
representations can be difficult to relate directly to physically meaningful 
propagation patterns.

\subsection{Research Gap}

Existing graph-classification methods offer complementary strengths but leave an 
important gap for cyber-physical graph mining. Kernel methods can be expensive for 
large graph collections, generic graph descriptors may overlook domain-specific 
edge semantics, and GNN representations\cite{gcn,gat,gine,transformerconv} are learned through end-to-end training. 
RWF~\cite{rwf} provides an attractive middle ground because it is lightweight, 
interpretable, and scalable. However, its standard transition dynamics are 
primarily topology-driven, while optional attribute enrichment focuses on node 
features. In cyber-physical systems, edge attributes often encode critical 
interaction states. For example, in a cyber-physical power grid, flow intensity, capacity margin, coupling strength, 
and overload severity can be represented using edge weights. These attributes should influence how random walks 
propagate across the network. To address this limitation, the proposed MC-PA-RWF 
constructs multiple physics-aware edge-weighted channels, extracts a 
channel-specific fingerprint from each transition process, and concatenates the 
resulting fingerprints into a unified graph-level representation.

\section{Preliminaries and Background}
\label{sec:preliminaries}

\subsection{Attributed Graph Classification}
\label{subsec:attributed_graph_classification}

Let $\mathcal{G}=\{G_1,\ldots,G_N\}$ denote a collection of attributed graphs,
where each graph is represented as
\[
G_i=(V_i,E_i,A_i,X_i,H_i,y_i).
\]
Here, $V_i$ and $E_i$ are the node and edge sets,
$A_i\in\mathbb{R}^{|V_i|\times |V_i|}$ is the topology-based adjacency matrix,
$X_i\in\mathbb{R}^{|V_i|\times p}$ is the node-feature matrix,
$H_i\in\mathbb{R}^{|E_i|\times q}$ is the edge-feature matrix, and $y_i$ is
the graph-level label.

The objective is to learn a mapping
\[
f:G_i\rightarrow y_i.
\]
We follow a feature-extraction approach in which each graph is transformed into
a fixed-length vector,
\[
\phi(G_i)=z_i\in\mathbb{R}^{d},
\]
and a classifier is trained on the resulting representations. For
cyber-physical graphs, $\phi(\cdot)$ should capture not only the structural
connectivity encoded by $A_i$, but also the operational interaction patterns
encoded by $H_i$.

\subsection{Random Walk Fingerprints (RWF)}
\label{subsec:rwf}

RWF construct an unsupervised graph-level
representation by summarising random-walk connectivity over structurally aligned
node groups~\cite{rwf}. Given an adjacency matrix $A$, RWF defines the
symmetrically normalised walk (propagation) matrix
\[
T=D^{-\frac{1}{2}}AD^{-\frac{1}{2}},
\qquad
D_{uu}=\sum_v A_{uv},
\]
where $(T^\tau)_{uv}$ captures the strength of $\tau$-step walks from node $u$
to node $v$.

RWF partitions the node set into $k$ structurally aligned groups,
\[
V=V_1\cup\cdots\cup V_k,
\]
and computes loop, within-group, and cross-group summaries for
$\tau\in\{1,\ldots,\tau_c\}$. For node groups $V_a$ and $V_b$,
\[
\begin{aligned}
\mathrm{loop}^{\tau}_{V_a}
&=
\frac{1}{|V_a|}
\sum_{u\in V_a}(T^\tau)_{uu},\\
\mathrm{walk}^{\tau}_{V_a}
&=
\frac{1}{|V_a|^2}
\sum_{u,v\in V_a}(T^\tau)_{uv},\\
\mathrm{walk}^{\tau}_{V_a\rightarrow V_b}
&=
\frac{1}{|V_a||V_b|}
\sum_{\substack{u\in V_a\\v\in V_b}}
(T^\tau)_{uv}.
\end{aligned}
\]
Collecting these statistics across walk lengths and group pairs yields a
fixed-length fingerprint
\[
z_G=\phi_{\mathrm{RWF}}(G)\in\mathbb{R}^{d}.
\]
The representation is invariant to node ordering and compatible with graphs of
different sizes. RWF can also incorporate node-feature interactions, but its
transition process is primarily driven by topology~\cite{rwf}.

\subsection{Motivation for Physics-Aware Edge Channels}
\label{subsec:edge_attribute_motivation}

Topology alone is insufficient when graph-level outcomes depend on edge states.
Let $h_{uv}\in\mathbb{R}^{q}$ denote the feature vector of edge $(u,v)\in E$.
A binary adjacency matrix records whether two nodes are connected, but it does
not distinguish between lightly loaded and stressed interactions. In a power
grid, for example, a line carrying high power flow with limited residual capacity
should influence corresponding graph walk propagation differently from a lightly loaded line with a
large remaining capacity margin. In simple terms, power flow variations across lines will influence the respective graph walk dynamics differently. This motivates an edge-attributed extension of RWF in which random-walk
transitions are guided by multiple physical edge channels rather than by a
single topology-based adjacency matrix. The proposed MC-PA-RWF construction is
introduced in Section~\ref{sec:method}.

\section{Physics-Aware Multi-Channel Random Walk Fingerprints}
\label{sec:method}

This section presents Multi-Channel Physics-Aware Random Walk Fingerprints
(MC-PA-RWF), a graph-level representation for attributed cyber-physical
systems. Classical RWF summarises walk patterns using a single topology-based
transition matrix. In contrast, MC-PA-RWF constructs multiple physics-aware
edge-weighted channels from edge attributes and extracts an RWF block from each
channel. The resulting blocks are concatenated into a fixed-length graph
representation. We also introduce MC-PA-RWF+, a node-edge extension that
combines selected edge-channel fingerprints with an RWF-D-feature block, presented in Section~\ref{subsec:mc_pa_rwf_plus}.

\subsection{Problem Formulation and Overview}
\label{subsec:method_overview}

Consider an attributed graph
\[
G=(V,E,A,X,H,y)
\]
where $V$ and $E$ are the node and edge sets,
$A$ is the topological adjacency matrix,
$X\in\mathbb{R}^{|V|\times p}$ contains node attributes,
$H\in\mathbb{R}^{|E|\times q}$ contains edge attributes, and $y$ is the
graph-level label. The goal is to construct a fixed-length fingerprint $z_G$
and train a classifier
\[
\hat{y}=f_{\theta}(z_G).
\]

MC-PA-RWF transforms the edge attributes into a set of weighted adjacency
matrices,
\[
\mathcal{A}_G=\{A^{(c)}:c\in\mathcal{C}\},
\]
where each channel $c$ represents a distinct physical interaction pattern over
the same graph support. A structural RWF fingerprint is extracted independently
from each channel, and the final representation is
\[
z_G^{\mathrm{MC}}
=
\big\Vert_{c\in\mathcal{C}} z_G^{(c)},
\]
where $\Vert$ denotes concatenation. Thus, MC-PA-RWF represents a graph through
multiple edge-state-dependent propagation views rather than a single
topology-based transition process.

\subsection{Physics-Aware Edge-Channel Construction}
\label{subsec:edge_channel_construction}

MC-PA-RWF represents a cyber-physical graph through multiple edge-weighted
views of the same topology. Each view emphasises a different physical property
of the interaction between connected components. For an edge $(u,v)\in E$,
let $h_{uv}$ denote its edge-feature vector. A channel map $\psi_c(\cdot)$
converts $h_{uv}$ into a non-negative undirected weight:
\[
A^{(c)}_{uv}=A^{(c)}_{vu}=
\begin{cases}
\psi_c(h_{uv}), & (u,v)\in E,\\
0, & \text{otherwise}.
\end{cases}
\]
Thus, the topology is preserved, while the channel $c$ modifies the walk dynamics.

For the power-grid instantiation, let
$h_{uv}=(P_{uv},Q_{uv},x_{uv},r_{uv})$, where $P_{uv}$ and $Q_{uv}$ are
active and reactive power flows, $x_{uv}$ is line reactance, and $r_{uv}$ is
the line rating. Since the proposed walk matrices are symmetric, signed or
directional quantities are converted to magnitudes. We define apparent flow
$S_{uv}=\sqrt{|P_{uv}|^2+|Q_{uv}|^2}$, loading ratio
$\rho_{uv}=S_{uv}/|r_{uv}|$, overload
$o_{uv}=\max(0,\rho_{uv}-1)$, remaining margin
$m_{uv}=\max(0,|r_{uv}|-S_{uv})$, and coupling proxy
$b_{uv}=1/(|x_{uv}|+\epsilon)$.

These quantities capture complementary operating conditions. High flow
indicates heavily utilised transmission paths; a large loading ratio or small
remaining margin indicates limited spare capacity; reactance-derived quantities
reflect the electrical strength of a connection; and overload-related measures
identify stressed lines that may contribute to cascading failures. The scalar
channels are organised into four groups:
\[
\begin{array}{ll}
\mathrm{FI}:\{|P_{uv}|,|Q_{uv}|,S_{uv}\}, &
\mathrm{CM}:\{\rho_{uv},|r_{uv}|^{-1}\},\\
\mathrm{CS}:\{|x_{uv}|,b_{uv}\}, &
\mathrm{OS}:\{\rho_{uv},o_{uv},(m_{uv}+\epsilon)^{-1}\}.
\end{array}
\]
Here, FI captures power-transfer intensity, CM capacity usage, CS electrical
coupling, and OS overload-related stress. Each scalar quantity defines a
separate weighted view and contributes a channel-specific RWF block.

To reduce scale differences and limit extreme values, each scalar channel is
median-scaled and clipped as
$A^{(c)}_{uv}
=
1+\operatorname{clip}
\left(
s^{(c)}_{uv}/\operatorname{med}_{+}(s^{(c)}),\,0,\,10
\right)$,
where $\operatorname{med}_{+}(s^{(c)})$ is the median of positive values; if
none exist, the normalised term is set to zero. The additive constant preserves
edge support.

\subsection{Channel-Wise Random Walk Fingerprints}
\label{subsec:channel_fingerprint_fusion}
Present MC-PA-RWF implementation constructs undirected weighted channels. 
Hence, signed or directional quantities are converted to non-negative magnitudes, and
each channel weight is assigned symmetrically, so
$A^{(c)}*{uv}=A^{(c)}*{vu}$. For each channel $c$, this yields the symmetrically normalised
walk (propagation) matrix 
\[
T^{(c)}
=
{D^{(c)}}^{-1/2}
A^{(c)}
{D^{(c)}}^{-1/2},
\qquad
D^{(c)}_{uu}
=
\sum_v A^{(c)}_{uv}.
\]
For zero-degree nodes, the corresponding entries of
${D^{(c)}}^{-1/2}$ are set to zero. For a walk length
$\tau\in\{1,\ldots,\tau_c\}$,
$\left(T^{(c)}\right)^\tau$
describes $\tau$-step propagation under the physical interpretation of channel
$c$. For example, FI emphasises highly loaded paths, whereas OS emphasises
stressed or overloaded lines.

Following RWF, nodes are partitioned into $k$ structural groups. In our
implementation, each channel induces its own degree-based ordering from the
corresponding weighted graph. Let
\[
\mathcal{P}^{(c)}
=
\{V_1^{(c)},\ldots,V_k^{(c)}\}
\]
denote the resulting partition. For a group $V_a^{(c)}$, the loop and
within-group summaries are
\[
\ell_{a}^{\tau,c}
=
\frac{1}{|V_a^{(c)}|}
\sum_{u\in V_a^{(c)}}
\left[\left(T^{(c)}\right)^\tau\right]_{uu},
\]
and
\[
w_{a}^{\tau,c}
=
\frac{1}{|V_a^{(c)}|^2}
\sum_{u,v\in V_a^{(c)}}
\left[\left(T^{(c)}\right)^\tau\right]_{uv}.
\]
For two distinct groups $V_a^{(c)}$ and $V_b^{(c)}$, the between-group
summary is
\[
w_{a,b}^{\tau,c}
=
\frac{1}{|V_a^{(c)}||V_b^{(c)}|}
\sum_{\substack{u\in V_a^{(c)}\\v\in V_b^{(c)}}}
\left[\left(T^{(c)}\right)^\tau\right]_{uv}.
\]

The channel fingerprint
\[
z_G^{(c)}
=
\phi_{\mathrm{RWF}}
\left(T^{(c)},\mathcal{P}^{(c)};\tau_c\right)
\]
is obtained by collecting these summaries across all groups and walk lengths.
Concatenating the channel fingerprints yields $z_G^{\mathrm{MC}}$. Unlike
topology-only RWF, MC-PA-RWF can distinguish graphs that share the same
connectivity but differ in their physical edge states.

\subsection{MC-PA-RWF+ Node-Edge Extension}
\label{subsec:mc_pa_rwf_plus}

MC-PA-RWF models edge-channel propagation but does not include node-feature
interactions inside the channel-wise extractor. MC-PA-RWF+ incorporates
node-level operating information by appending an RWF-D-feature block to selected
physics-aware edge-channel fingerprints:
\[
z_G^{\mathrm{MC+}}
=
z_G^{D\text{-feat}}
\Vert
\left(
\big\Vert_{c\in\mathcal{C}_{+}}z_G^{(c)}
\right),
\qquad
\mathcal{C}_{+}\subseteq\mathcal{C}.
\]

For MC-PA-RWF+, the channel subset selected on the validation split is
$\mathcal{C}_{+}=\{\mathrm{FI},\mathrm{CM},\mathrm{OS}\}$, yielding
\[
z_G^{\mathrm{MC+}}
=
z_G^{D\text{-feat}}
\Vert z_G^{(\mathrm{FI})}
\Vert z_G^{(\mathrm{CM})}
\Vert z_G^{(\mathrm{OS})}.
\]

The node-feature block $z_G^{D\text{-feat}}$ is computed once using the
RWF-D-feature mechanism~\cite{rwf}. The selected edge-channel blocks contribute
structural summaries over their weighted walk (propagation) matrices. This modular
design combines local node-feature evidence with edge-state
information without duplicating node-feature interactions across channels.

Both MC-PA-RWF and MC-PA-RWF+ produce fixed-length graph-level vectors. In this
work, an SVM is trained on the extracted fingerprints, separating
physics-aware feature extraction from supervised classification.

\begin{table}[t]
\centering
\caption{Time complexity comparison of graph methods.}
\label{tab:complexity_comparison}
\setlength{\tabcolsep}{4pt}
\renewcommand{\arraystretch}{1.08}
\begin{tabular}{ll}
\toprule
Method & Time complexity \\
\midrule
Random walk kernel~\cite{random_walk_kernel} & $O(N^2n^3)$ \\
NetLSD~\cite{netlsd} & $O(Nrm + Nr^2n)$ \\
FGSD~\cite{fgsd} & $O(Nn^2)$ \\
NetSimile~\cite{netsimile} & $O(Nf(n+n\log n))$ \\
\midrule
RWF-D~\cite{rwf} & $O(N\tau_c mn)$ \\
\shortstack[l]{RWF-D-feature\cite{rwf}} & $O(N\tau_c(mn+pn^2))$ \\
MC-PA-RWF (Proposed) & $O(NC\tau_c mn)$ \\
MC-PA-RWF+ (Proposed) & $O(N\tau_c((C_+ + 1)mn + pn^2))$ \\
\bottomrule
\end{tabular}

\vspace{1mm}
\footnotesize{
\emph{Note:}
Here, $N$ denotes the number of graphs, $r$ the number of eigenvalues retained for the NetLSD approximation, and $f$ the number of node-level structural features used by NetSimile.
}
\vspace{-1em}
\end{table}

\subsection{Complexity Analysis}
\label{subsec:complexity_analysis}

Table~\ref{tab:complexity_comparison} compares the feature-extraction costs of
the proposed methods with representative graph-representation approaches. Let
$N$ denote the number of graphs, $n$ and $m$ the maximum numbers of nodes and
edges per graph, $\tau_c$ the maximum walk length, $p$ the node-feature
dimension, $C=|\mathcal{C}|$ the number of physical channels, and
$C_{+}=|\mathcal{C}_{+}|$ the number of channels used by MC-PA-RWF+.

Under the sparse RWF assumptions of~\cite{rwf}, topology-only RWF-D has
complexity $O(N\tau_c mn)$, while RWF-D-feature requires
$O\!\left(N\tau_c(mn+pn^2)\right)$ due to additional node-feature
interactions. MC-PA-RWF applies structural RWF extraction independently to
$C$ edge-weighted symmetrically normalised walk matrices, resulting in $O(NC\tau_c mn)$. Constructing
the weighted adjacency matrices costs $O(NCm)$, while channel-wise
degree-based partitioning costs $O\!\left(NC(m+n\log n)\right)$. These
are lower order than repeated fingerprint extraction.

MC-PA-RWF+ combines one RWF-D-feature block with $C_{+}$ selected
edge-channel blocks, giving
$O\!\left(N\tau_c\left((C_{+}+1)mn+pn^2\right)\right)$. Both
proposed variants remain linear in the number of graph instances $N$. Their
dependence on the size of each graph is inherited from the underlying RWF
computation, while the additional overhead is controlled by the small, fixed
number of physical channels.

\subsection{Algorithm}
\label{subsec:algorithm}

Algorithm~\ref{alg:mc_pa_rwf} summarises both MC-PA-RWF and MC-PA-RWF+. Both variants extract structural RWF fingerprints from physics-aware edge-weighted channels. MC-PA-RWF uses all physical channels in $\mathcal{C}$ and concatenates only the resulting edge-channel fingerprints. MC-PA-RWF+ additionally includes an RWF-D-feature block $z_{G_i}^{D\text{-feat}}$, computed from topology and node attributes, and then appends selected edge-channel fingerprints from $\mathcal{C}_+$. Thus, node-feature interactions are included only through the RWF-D-feature block, while the physical channels use structural RWF summaries over weighted walk matrices.

\begin{algorithm}[t]
\caption{MC-PA-RWF and MC-PA-RWF+ Feature Extraction}
\label{alg:mc_pa_rwf}
\KwIn{Graph dataset $\mathcal{G}=\{G_i\}_{i=1}^{N}$, where $G_i=(V_i,E_i,A_i,X_i,H_i,y_i)$; physical channel set $\mathcal{C}$; selected channel set $\mathcal{C}_+\subseteq\mathcal{C}$ for MC-PA-RWF+; edge-channel maps $\{\psi_c\}_{c\in\mathcal{C}}$; number of node groups $k$; maximum walk length $\tau_c$; mode $\eta\in\{\mathrm{MC\text{-}PA\text{-}RWF},\mathrm{MC\text{-}PA\text{-}RWF+}\}$.}
\KwOut{Graph fingerprints $\{z_{G_i}\}_{i=1}^{N}$.}

\ForEach{$G_i\in\mathcal{G}$}{
    Initialise $\mathcal{Z}_i\leftarrow[\,]$\;

    \If{$\eta=\mathrm{MC\text{-}PA\text{-}RWF+}$}{
        Compute node-feature RWF block
        $z_{G_i}^{D\text{-feat}}=\mathrm{RWF}_{\mathrm{feat}}(A_i,X_i;k,\tau_c)$\;
        Append $z_{G_i}^{D\text{-feat}}$ to $\mathcal{Z}_i$\;
        Set active channel set $\mathcal{C}^{\star}\leftarrow\mathcal{C}_+$\;
    }
    \Else{
        Set active channel set $\mathcal{C}^{\star}\leftarrow\mathcal{C}$\;
    }

    \ForEach{$c\in\mathcal{C}^{\star}$}{
        Construct physical channel adjacency
        $A_{i,uv}^{(c)}=\psi_c(h_{i,uv})$ for $(u,v)\in E_i$\;
        Normalise $A_i^{(c)}$ as a non-negative weighted graph\;
        Partition nodes into $k$ RWF groups $\mathcal{P}_i^{(c)}$ using channel-weighted degrees\;
        Compute $T_i^{(c)}={D_i^{(c)}}^{-1/2}A_i^{(c)}{D_i^{(c)}}^{-1/2}$\;
        Extract structural RWF summaries
        $z_{G_i}^{(c)}=\mathrm{RWF}_{\mathrm{str}}(T_i^{(c)},\mathcal{P}_i^{(c)};\tau_c)$\;
        Append $z_{G_i}^{(c)}$ to $\mathcal{Z}_i$\;
    }

    Concatenate selected blocks:
    $z_{G_i}=\Vert_{z\in\mathcal{Z}_i}z$\;
}
\Return{$\{z_{G_i}\}_{i=1}^{N}$}\;
\end{algorithm}

\begin{table}
\centering
\caption{PowerGraph graph-level classification datasets~\cite{powergraph}.}
\label{tab:datasets}
\setlength{\tabcolsep}{3.5pt}
\renewcommand{\arraystretch}{1.15}
\begin{tabularx}{\columnwidth}{l c c c c X}
\hline
\textbf{Dataset} & \textbf{Nodes} & \textbf{Edges} & \textbf{Graphs} & \textbf{Used} & \textbf{Classes} \\
\hline
IEEE24  & 24  & 38  & 21,500  & 1k--21.5k   & Stable / Failure-prone \\
IEEE39  & 39  & 46  & 28,000  & 1k--28k     & Stable / Failure-prone \\
UK      & 29  & 99  & 64,000  & 5k--64k     & Stable / Failure-prone \\
\hline
\end{tabularx}
\end{table}

\section{Experimental Setup}
\label{sec:experimental_setup}

This section describes the datasets, baseline methods used for comparison, and the evaluation metrics used to assess MC-PA-RWF and MC-PA-RWF+. The experiments evaluate graph-level cascading-failure classification on PowerGraph benchmark systems under increasing dataset sizes. We compare the proposed physics-aware random-walk fingerprints with original RWF variants and supervised GNN baselines, using balanced accuracy as the primary metric and failure-class F1 to assess how reliably each method identifies failure-prone operating scenarios.

\begin{table*}[t]
\centering
\caption{IEEE24 bus test system balanced accuracy (\%) across different graph sizes}
\label{tab:ieee24_results}
\setlength{\tabcolsep}{3.5pt}
\begin{tabular}{llrrrrrr}
\toprule
Family & Method & 1k & 3k & 5k & 10k & 15k & 21.5k \\
\midrule
RWF (Original) & RWF-D~\cite{rwf} & 71.78 $\pm$ 4.15 & 72.29 $\pm$ 4.12 & 66.75 $\pm$ 1.22 & 67.51 $\pm$ 1.77 & 67.06 $\pm$ 1.26 & 67.04 $\pm$ 0.74 \\
 & RWF-D-feature~\cite{rwf} & 93.00 $\pm$ 5.70 & 96.09 $\pm$ 1.86 & 96.33 $\pm$ 1.59 & 97.02 $\pm$ 1.11 & 97.50 $\pm$ 1.00 & 97.81 $\pm$ 0.45 \\
\midrule
GNN & GCN~\cite{gcn} & 75.53 $\pm$ 2.66 & 85.01 $\pm$ 4.50 & 89.04 $\pm$ 1.54 & 92.80 $\pm$ 0.85 & 93.90 $\pm$ 1.14 & 95.23 $\pm$ 1.25 \\
 & GAT~\cite{gat} & 76.98 $\pm$ 2.06 & 88.50 $\pm$ 1.95 & 90.29 $\pm$ 1.38 & 91.96 $\pm$ 1.47 & 93.34 $\pm$ 1.99 & 92.90 $\pm$ 2.06 \\
 & GINE~\cite{gine} & 83.38 $\pm$ 4.54 & 93.45 $\pm$ 2.74 & 92.51 $\pm$ 2.81 & 96.82 $\pm$ 0.62 & 97.78 $\pm$ 0.51 & 97.84 $\pm$ 0.82 \\
 & TransformerConv~\cite{transformerconv} & 88.43 $\pm$ 3.20 & 94.42 $\pm$ 1.96 & 95.23 $\pm$ 1.93 & 97.59 $\pm$ 0.67 & 98.06 $\pm$ 1.08 & 98.76 $\pm$ 0.49 \\
\midrule
\multirow[t]{6}{2.6cm}{\raggedright Physics-aware RWF Variants\\} & PA-RWF-FI & 92.45 $\pm$ 3.93 & 96.79 $\pm$ 1.44 & 96.38 $\pm$ 1.14 & 97.37 $\pm$ 0.70 & 98.53 $\pm$ 0.64 & 98.77 $\pm$ 0.48 \\
 & PA-RWF-CM & 91.15 $\pm$ 5.08 & 96.10 $\pm$ 1.12 & 96.63 $\pm$ 1.67 & 98.09 $\pm$ 0.69 & 98.63 $\pm$ 0.49 & 98.95 $\pm$ 0.33 \\
 & PA-RWF-CS & 70.58 $\pm$ 4.78 & 72.17 $\pm$ 3.94 & 66.93 $\pm$ 2.75 & 67.04 $\pm$ 1.70 & 66.72 $\pm$ 1.65 & 67.62 $\pm$ 0.91 \\
 & PA-RWF-OS & \textbf{94.85 $\pm$ 2.74} & 96.42 $\pm$ 1.25 & 96.48 $\pm$ 1.59 & 98.12 $\pm$ 0.43 & 98.65 $\pm$ 0.28 & 98.60 $\pm$ 0.35 \\
 & MC-PA-RWF (Proposed) & 94.51 $\pm$ 4.23 & \textbf{97.08 $\pm$ 0.95} & \textbf{97.68 $\pm$ 1.68} & \textbf{98.43 $\pm$ 0.29} & 98.71 $\pm$ 0.50 & 99.15 $\pm$ 0.22 \\
 & MC-PA-RWF+ (Proposed) & 93.54 $\pm$ 6.90 & 96.47 $\pm$ 1.27 & 97.48 $\pm$ 1.41 & 98.17 $\pm$ 0.16 & \textbf{99.00 $\pm$ 0.30} & \textbf{99.21 $\pm$ 0.36} \\
\bottomrule
\end{tabular}
\end{table*}

\subsection{Dataset and Task}
\label{subsec:dataset_task}

We evaluate the proposed method on the graph-level cascading-failure classification benchmark from \textit{PowerGraph}~\cite{powergraph}\footnote{\url{https://figshare.com/articles/dataset/PowerGraph/22820534}}. {PowerGraph} provides attributed graph datasets for cyber-physical energy-system learning tasks, including power-flow analysis, optimal-power-flow analysis, and cascading-failure prediction. In the graph-level cascading-failure setting, each graph represents one operating scenario of a power-grid system associated with an initial triggering outage. The graph-level label is derived from the final demand not served (DNS) produced by the physics-based cascading-failure simulation.

In this work, we use the binary classification setting. A graph is labelled as stable if the cascading-failure simulation results in no demand not served, and failure-prone otherwise:
\[
y_i =
\begin{cases}
0, & \text{if } \mathrm{DNS}_i = 0,\\
1, & \text{if } \mathrm{DNS}_i > 0.
\end{cases}
\]
Here, $y_i=0$ denotes Stable/Class 0 and $y_i=1$ denotes Failure-prone/Class 1. Each graph is represented as
\[
G_i=(V_i,E_i,A_i,X_i,H_i,y_i),
\]
where nodes correspond to grid components, edges correspond to physical connections, $A_i$ is the topology-based adjacency matrix, $X_i$ denotes node attributes, and $H_i$ denotes edge attributes. The proposed MC-PA-RWF uses the topology and edge attributes to construct physical edge-weighted channels representing complementary line semantics, including flow intensity, capacity margin, coupling strength, and overload severity. Node attributes are not used inside the MC-PA-RWF edge-channel extractor. They are used in RWF-D-feature and in the proposed MC-PA-RWF+ extension, where an RWF-D-feature block computed from topology and node attributes is appended to selected physics-aware edge-channel fingerprints. Thus, MC-PA-RWF evaluates edge-attributed physical propagation, while MC-PA-RWF+ combines node-feature evidence with physics-aware edge-channel random-walk summaries. We use only input graph attributes and graph-level labels for training and evaluation; ground-truth explanation files, such as failed-edge explanation labels, are not used as input features. Table~\ref{tab:datasets} summarises the benchmark systems and graph sizes used in the experiments.

\subsection{Benchmark Methods}
\label{subsec:compared_methods}

We compare MC-PA-RWF with three groups of methods.

\textbf{RWF baselines:}
We include two direct RWF baselines. \textbf{RWF-D} is the topology-only RWF variant using degree-based structural-role partitioning~\cite{rwf}. It computes random-walk fingerprints from the topology-based adjacency matrix only. \textbf{RWF-D-feature} extends RWF-D by incorporating node attributes through the node-feature aggregation mechanism of RWF~\cite{rwf}. These baselines allow us to evaluate whether edge-channel fingerprints provide additional discriminative information beyond topology and node attributes.

\textbf{Physics-aware RWF variants (Proposed):}
We evaluate two proposed physics-aware RWF variants. \textbf{MC-PA-RWF} constructs multiple physical edge-weighted graph channels and concatenates the structural RWF fingerprints extracted from these channels. This variant uses topology and edge attributes only. \textbf{MC-PA-RWF+} extends MC-PA-RWF by appending an RWF-D-feature block computed from topology and node attributes to selected physics-aware edge-channel fingerprints. Thus, MC-PA-RWF evaluates multi-channel edge-attributed random-walk propagation, while MC-PA-RWF+ evaluates the combined effect of node-feature evidence and physics-aware edge-channel propagation. Further details are provided in Section~\ref{sec:method}. 
To analyse the contribution of individual physical edge channels, we 
include four single-channel diagnostic variants: PA-RWF-FI, PA-RWF-CM,
PA-RWF-CS, and PA-RWF-OS. Each variant follows the MC-PA-RWF
fingerprint-extraction methodology but uses only one edge-weighted channel:
flow intensity, capacity margin, coupling strength, or overload severity,
respectively. These variants are included for channel-level analysis of the 
proposed models.

\textbf{Neural graph baselines:}
We compare against four supervised GNN baselines for graph-level classification: GCN~\cite{gcn}, GAT~\cite{gat}, GINE~\cite{gine}, and TransformerConv~\cite{transformerconv}. GCN provides a standard message-passing baseline based on neighbourhood feature aggregation, while GAT uses attention weights to learn the relative importance of neighbouring nodes. GINE extends the expressive GIN architecture by incorporating edge attributes into the message-passing update, making it a strong baseline for attributed power-grid graphs. TransformerConv uses attention-based graph convolution and provides a more flexible neural baseline for modelling graph interactions. All GNNs use the same graph-level labels and are evaluated with the same performance metric as the RWF-based methods.

\begin{figure*}[t]
  \centering
  \includegraphics[width=0.9\textwidth]{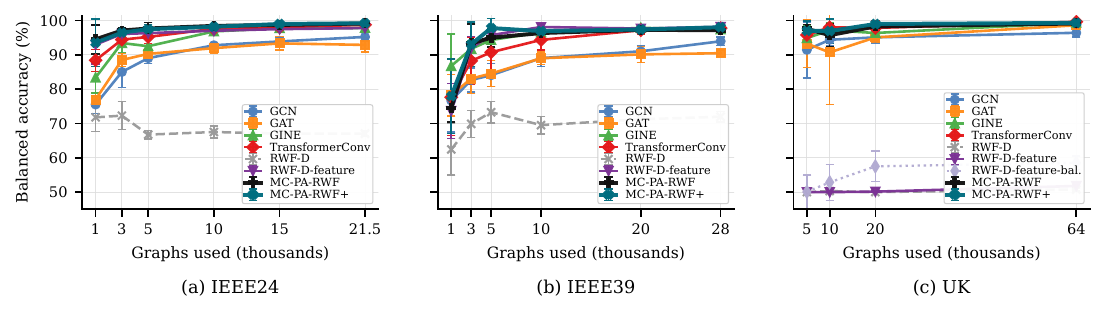}
  \caption{Classification performance across dataset sizes for IEEE24, IEEE39, and UK benchmark test systems.}
  \label{fig:benchmark_performance_three_panels}
\end{figure*}

\begin{table*}
\centering
\caption{IEEE39 bus test system balanced accuracy (\%) across different graph sizes.}
\label{tab:ieee39_size_saturation}
\setlength{\tabcolsep}{3.5pt}
\begin{tabular}{llrrrrrr}
\toprule
Family & Method & 1k & 3k & 5k & 10k & 20k & 28k \\
\midrule
RWF (Original) & RWF-D~\cite{rwf}  & 62.43 $\pm$ 7.51 & 69.88 $\pm$ 3.98 & 73.28 $\pm$ 3.16 & 69.51 $\pm$ 2.54 & 71.23 $\pm$ 1.86 & 71.96 $\pm$ 1.48 \\
 & RWF-D-feature~\cite{rwf}  & 73.65 $\pm$ 7.96 & 92.03 $\pm$ 3.35 & 95.74 $\pm$ 2.70 & \textbf{98.12 $\pm$ 0.83} & 97.66 $\pm$ 0.72 & \textbf{98.15 $\pm$ 0.37} \\
\midrule
GNN & GCN~\cite{gcn} & 76.67 $\pm$ 9.20 & 82.69 $\pm$ 3.50 & 84.07 $\pm$ 3.37 & 89.05 $\pm$ 2.50 & 91.04 $\pm$ 1.69 & 93.96 $\pm$ 1.30 \\
 & GAT~\cite{gat} & 78.22 $\pm$ 6.09 & 83.09 $\pm$ 4.33 & 84.56 $\pm$ 3.86 & 89.01 $\pm$ 1.86 & 90.13 $\pm$ 2.37 & 90.48 $\pm$ 0.96 \\
 & GINE~\cite{gine} & \textbf{86.75 $\pm$ 9.39} & 91.75 $\pm$ 2.49 & 94.33 $\pm$ 1.31 & 96.56 $\pm$ 1.52 & \textbf{97.87 $\pm$ 0.55} & 97.41 $\pm$ 0.80 \\
 & TransformerConv~\cite{transformerconv} & 77.63 $\pm$ 11.17 & 88.25 $\pm$ 6.94 & 90.76 $\pm$ 6.27 & 94.36 $\pm$ 2.98 & 97.17 $\pm$ 1.07 & 97.76 $\pm$ 0.81 \\
\midrule
\multirow[t]{2}{2.8cm}{\raggedright Physics-aware RWF\\(Proposed)}
 & MC-PA-RWF & 74.54 $\pm$ 4.11 & \textbf{93.32 $\pm$ 5.69} & 94.99 $\pm$ 2.76 & 96.35 $\pm$ 0.99 & 97.30 $\pm$ 1.02$^\dagger$ & 97.15 $\pm$ 0.48 \\
 & MC-PA-RWF+ & 77.93 $\pm$ 10.89 & 93.08 $\pm$ 6.53 & \textbf{97.80 $\pm$ 2.78} & 96.92 $\pm$ 1.19$^\dagger$ & 97.39 $\pm$ 0.89$^\dagger$ & 98.04 $\pm$ 0.54$^\dagger$ \\
\bottomrule
\end{tabular}

\vspace{1mm}
\begin{flushleft}
\footnotesize
\textit{Note:} Bold values indicate the highest mean. $\dagger$: not significantly different from the best method in the column under a paired two-sided $t$-test ($\alpha=0.05$).
\end{flushleft}
\end{table*}

\begin{table}[!t]
\centering
\caption{IEEE39 fusion results: balanced accuracy (\%)}
\label{tab:ieee39_adaptive_fusion_ablation}
\setlength{\tabcolsep}{2.5pt}
\renewcommand{\arraystretch}{1.15}
\small
\begin{tabular}{lcccccc}
\toprule
Method & 1k & 3k & 5k & 10k & 20k & 28k \\
\midrule
\shortstack[l]{RWF-D-feature \\\cite{rwf}\vphantom{$\pm$}} & \shortstack{82.05\\$\pm$ 6.76} & \shortstack{93.99\\$\pm$ 3.71} & \shortstack{97.22\\$\pm$ 1.70} & \shortstack{98.44\\$\pm$ 0.48} & \shortstack{98.51\\$\pm$ 0.34} & \shortstack{98.63\\$\pm$ 0.38} \\
\midrule
\shortstack[l]{MC-PA-RWF\\(Adpt.)} & \shortstack{\textbf{87.19}\\$\pm$ 7.16} & \shortstack{\textbf{95.46}\\$\pm$ 3.66} & \shortstack{\textbf{97.84}\\$\pm$ 2.57} & \shortstack{\textbf{98.50}\\$\pm$ 0.51} & \shortstack{\textbf{98.92}\\$\pm$ 0.39} & \shortstack{\textbf{98.91}\\$\pm$ 0.40} \\
\bottomrule
\end{tabular}
\normalsize
\renewcommand{\arraystretch}{1.0}
\end{table}

\begin{table*}[!t]
\centering
\caption{UK test system balanced accuracy (\%) across different graph sizes.}
\label{tab:uk_size_saturation_gnn_full}
\setlength{\tabcolsep}{4.5pt}
\begin{tabular}{llrrrr}
\toprule
Family & Method & 5k & 10k & 20k & 64k \\
\midrule
RWF (Original) & RWF-D~\cite{rwf} 
& 50.00 $\pm$ 0.00
& 50.30 $\pm$ 0.68
& 50.00 $\pm$ 0.00
& 50.77 $\pm$ 0.47 \\
& RWF-D-feature~\cite{rwf} 
& 50.00 $\pm$ 0.00
& 50.00 $\pm$ 0.00
& 50.14 $\pm$ 0.30
& 51.79 $\pm$ 0.27 \\
& RWF-D-feature-balanced
& 50.10 $\pm$ 4.85
& 52.84 $\pm$ 5.25
& 57.51 $\pm$ 4.41
& 58.54 $\pm$ 1.94 \\
\midrule
GNN & GCN~\cite{gcn}
& 91.46 $\pm$ 8.17
& 94.39 $\pm$ 1.75
& 95.10 $\pm$ 1.70
& 96.43 $\pm$ 1.25 \\
& GAT~\cite{gat}
& 93.28 $\pm$ 7.01
& 90.71 $\pm$ 15.25
& 95.09 $\pm$ 1.30
& 98.62 $\pm$ 0.61 \\
& GINE~\cite{gine}
& 94.87 $\pm$ 3.72
& 97.16 $\pm$ 2.16
& 96.37 $\pm$ 1.87
& 99.36 $\pm$ 0.30 \\
& TransformerConv~\cite{transformerconv}
& 95.79 $\pm$ 4.19
& \textbf{98.01 $\pm$ 1.13}
& 97.85 $\pm$ 2.16
& \textbf{99.63 $\pm$ 0.43} \\
\midrule
\multirow[t]{2}{2.8cm}{\raggedright Physics-aware RWF\\(Proposed)}
& MC-PA-RWF
& \textbf{97.59 $\pm$ 1.24}
& 95.74 $\pm$ 3.28$^{\dagger}$
& 98.27 $\pm$ 1.09$^{\dagger}$
& 99.14 $\pm$ 0.59$^{\dagger}$ \\
& MC-PA-RWF+
& 97.00 $\pm$ 2.86$^{\dagger}$
& 96.94 $\pm$ 3.56$^{\dagger}$
& \textbf{98.99 $\pm$ 0.38}
& 99.32 $\pm$ 0.35$^{\dagger}$ \\
\bottomrule
\end{tabular}

\vspace{1mm}
\footnotesize{\emph{Note:} Bold values indicate the highest mean. $\dagger$: not significantly different from the best method in the column under a paired two-sided $t$-test ($\alpha=0.05$).
}
\end{table*}

\begin{table}[t]
\centering
\caption{Failure Class comparison against the best GNNs}
\label{tab:failure_f1_gnn_sig}
\setlength{\tabcolsep}{3.2pt}
\renewcommand{\arraystretch}{1.12}
\begin{tabular}{lclcccc}
\toprule
Dataset & Size & Best GNN & GNN F1 & Prop. F1 & $p$-value & Stat. Sig. \\
\midrule
IEEE24~\cite{powergraph} & 21.5k & TransConv
& 97.24\% & \textbf{98.84\%} & 0.0015 & \textbf{Yes} \\
IEEE39~\cite{powergraph} & 28k & TransConv
& 91.03\% & \textbf{96.87\%} & 0.0024 & \textbf{Yes} \\
UK~\cite{powergraph} & 64k & TransConv
& 96.95\% & \textbf{99.05\%} & 0.0147 & \textbf{Yes} \\
\bottomrule
\end{tabular}

\vspace{1mm}
\footnotesize{\emph{Note:} Prop. F1: MC-PA-RWF+ failure-class F1. Stat. Sig.: significant gain over the best GNN under a paired two-sided $t$-test ($p<0.05$).
}
\vspace{-1.2em}
\end{table}

\subsection{Evaluation Metrics and Implementation}
\label{subsec:evaluation_protocol}

Balanced accuracy is used as the primary metric because the stable and
failure-prone classes are imbalanced. Balanced accuracy is defined as
$\mathrm{BAcc}
=
\frac{1}{2}
\left(
\frac{\mathrm{TP}}{\mathrm{TP}+\mathrm{FN}}
+
\frac{\mathrm{TN}}{\mathrm{TN}+\mathrm{FP}}
\right)$.

We report the mean and standard deviation across repeated random seeds. At the
largest evaluated size of each benchmark, we additionally report failure-class
F1 to assess detection of the safety-critical failure-prone class.

For each split, preprocessing and model selection are performed using training
and validation data only. For the RWF-based methods, graph fingerprints are
first extracted and then classified using an SVM. SVM is widely used for classification tasks~\cite{anwar2015datadriven}. The number of node groups
$k$, maximum walk length $\tau_c$, SVM hyperparameters, and GNN
hyperparameters are selected from fixed search spaces using the validation
split. Feature standardisation is fitted on the training split and applied
unchanged to the validation and test splits. The test split is used only for
final evaluation.

The GNN baselines (GCN~\cite{gcn}, GAT~\cite{gat}, GINE~\cite{gine}, and TransformerConv~\cite{transformerconv}) are evaluated using
the same graph-level labels and balanced-accuracy metric. For reproduced
comparisons, the same train--validation--test splits and random seeds are used
across all methods. GNN training uses validation-based early stopping. Statistical comparisons are conducted across paired random seeds using a
two-sided paired $t$-test at the $0.05$ significance level.

\section{Results and Discussion}
\label{sec:results}
This section evaluates the proposed Physics-aware RWF framework (MC-PA-RWF and MC-PA-RWF+) through the following
research questions:

\begin{itemize}
    \item \textbf{RQ1:} Do physics-aware edge channels improve graph-level power grid
    cascading-failure classification compared with topology-only and
    node-feature RWF baselines?

    \item \textbf{RQ2:} Does combining complementary physical edge channels
    produce a more robust power grid graph representation than relying on a single
    channel or topology alone for such powerful RWF algorithms?

    \item \textbf{RQ3:} How do MC-PA-RWF and MC-PA-RWF+ compare with strong
    neural graph baselines (heavily used in power system) in terms of balanced accuracy and
    failure-class detection?
\end{itemize}

The results provide three main findings. (i) First, topology-only RWF suffers heavily for cascading-failure classification, particularly on the larger UK system. (ii) Second, operational edge channels such as flow intensity, capacity margin, and overload severity provide strong discriminative information, while their multi-channel combination improves robustness across dataset sizes. (iii) Finally, MC-PA-RWF+ is competitive with strong GNN baselines in balanced accuracy and significantly improves failure-class F1 on all three systems at the largest evaluated dataset sizes. In the results tables for this study, bold values indicate the highest mean balanced accuracy for each dataset size. A dagger ($\dagger$) indicates that the proposed method performs comparably to the best method in the same column, with no statistically significant difference under a paired two-sided $t$-test at the $0.05$ level.

\subsection{RQ1: Do Physics-Aware RWF Improve Classification?} \label{subsec:rq1} Tables~\ref{tab:ieee24_results}--\ref{tab:uk_size_saturation_gnn_full} and Fig.~\ref{fig:benchmark_performance_three_panels} show that topology alone does not provide a sufficiently informative representation for cascading-failure prediction. The topology-only RWF-D baseline reaches only $67.04\%$ balanced accuracy on IEEE24 at 21.5k graphs and $71.96\%$ on IEEE39 at 28k graphs. On the UK benchmark, RWF-D reaches only $50.77\%$ at 64k graphs, remaining
close to the $50\%$ balanced-accuracy baseline. This behaviour is expected because a binary adjacency matrix indicates whether two buses are connected but does not capture whether a transmission line is lightly loaded, close to its capacity limit, or part of a stressed propagation pathway. The node-feature baseline RWF-D-feature confirms that operating attributes are important. At the largest evaluated sizes, it improves balanced accuracy to $97.81\%$ on IEEE24 and $98.15\%$ on IEEE39. However, its performance is system-dependent. On the UK benchmark, RWF-D-feature achieves only $51.79\%$ balanced accuracy at 64k graphs. Even an additional diagnostic variant, RWF-D-feature-balanced, which uses the same RWF-D-feature representation but trains the SVM with class-balanced weights, reaches only $58.54\%$ at 64k graphs for the UK dataset of Table~\ref{tab:uk_size_saturation_gnn_full}. This contrast indicates that node-feature evidence alone does not generalise consistently across power-system settings. The proposed physics-aware representation addresses this limitation by allowing edge states to shape the random-walk transition dynamics. At the largest evaluated sizes, MC-PA-RWF reaches $99.15\%$ on IEEE24, $97.15\%$ on IEEE39, and $99.14\%$ on the UK system. The node-edge extension MC-PA-RWF+ reaches $99.21\%$, $98.04\%$, and $99.32\%$, respectively. The improvement is particularly clear on the UK benchmark, where the proposed methods remain highly accurate while the original RWF variants remain close to the 50\% balanced-accuracy baseline. These findings answer RQ1 positively: physics-aware edge channels provide substantial additional information beyond topology-only RWF and improve robustness when node-feature evidence alone is insufficient.

\subsection{RQ2: Does Multi-Channel Fusion Improve Robustness?}
\label{subsec:rq2}

To assess the contribution of each physical edge channel, we first evaluate
four single-channel diagnostic variants on the IEEE24 benchmark:
PA-RWF-FI, PA-RWF-CM, PA-RWF-CS, and PA-RWF-OS. At 21.5k graphs, PA-RWF-FI reaches $98.77\%$ balanced accuracy,
PA-RWF-CM reaches $98.95\%$, and PA-RWF-OS reaches $98.60\%$. In contrast,
PA-RWF-CS reaches only $67.62\%$, remaining close to the topology-only RWF-D
baseline. This difference is physically meaningful. Flow intensity, capacity margin, and
overload severity reflect the operating condition of transmission lines under a
scenario and are therefore closely related to how cascading failures spread
through the network. Coupling strength captures an important electrical property,
but it is comparatively static and less discriminative when used alone. The
single-channel analysis therefore provides interpretable evidence that the
channels contribute differently to cascading-failure classification.

The proposed MC-PA-RWF combines these complementary edge views into a unified
fingerprint. On IEEE24, MC-PA-RWF achieves the highest mean balanced accuracy
at 3k, 5k, and 10k graphs, reaching $97.08\%$, $97.68\%$, and $98.43\%$,
respectively. MC-PA-RWF+ becomes the strongest method at 15k and 21.5k graphs,
reaching $99.00\%$ and $99.21\%$. At 21.5k graphs, MC-PA-RWF+ improves over
the strongest single-channel variant, PA-RWF-CM, by $0.26$ percentage points.

As an additional diagnostic comparison, we report the IEEE39 score-level fusion variant denoted as MC-PA-RWF (Adpt.). This variant combines calibrated SVM outputs from the RWF-D-feature and the physics-aware edge-channel representation (MC-PA-RWF). It is not treated as a separate proposed method; rather, it is included to test whether node-feature and edge-channel fingerprints provide complementary decision evidence. 
Table~\ref{tab:ieee39_adaptive_fusion_ablation} reports this calibrated score-level fusion
experiment under its own ten-seed evaluation scenario (its values should therefore not be directly compared with those in Table~\ref{tab:ieee39_size_saturation}). Across the evaluation, MC-PA-RWF (Adpt.) improves over RWF-D-feature at every dataset size. Balanced accuracy increases from $82.05\%$ to $87.19\%$ at 1k graphs and from $98.63\%$ to $98.91\%$ at 28k graphs. The smaller gain at larger sizes is expected as performance approaches saturation.

These results answer RQ2 positively. No single edge channel is sufficient
across all settings. Combining complementary physical channels improves the
robustness of the graph representation, while the node-edge extension provides
additional benefit when both local node states and line-level propagation patterns are informative.

\subsection{RQ3: How Do the Proposed Methods Compare with GNN Baselines?} \label{subsec:rq3} The proposed methods are competitive with strong supervised GNN baselines across the three systems. At the largest IEEE24 setting, TransformerConv is the strongest GNN with $98.76\%$ balanced accuracy, whereas MC-PA-RWF+ reaches $99.21\%$, an improvement of $0.45$ percentage points. On IEEE39 at 28k graphs, TransformerConv reaches $97.76\%$, while MC-PA-RWF+ reaches $98.04\%$, an improvement of $0.28$ percentage points. On the larger UK benchmark at 64k graphs, TransformerConv achieves the highest mean balanced accuracy of $99.63\%$, while MC-PA-RWF+ remains close at $99.32\%$. The difference is not statistically significant under the paired test. The dataset-size results provide a more detailed view. On IEEE39, MC-PA-RWF is the best-performing method at 3k graphs with $93.32\%$, while MC-PA-RWF+ is best at 5k graphs with $97.80\%$. At 20k and 28k graphs, several proposed-method results are dagger-marked, indicating that their differences from the strongest method are not statistically significant. On the UK benchmark, MC-PA-RWF is best at 5k graphs with $97.59\%$, and MC-PA-RWF+ is best at 20k graphs with $98.99\%$. At 64k graphs, MC-PA-RWF+ remains statistically comparable to TransformerConv. These results show that a non-neural, physics-aware graph fingerprint can match or exceed strong message-passing architectures in several regimes while remaining more transparent. Unlike GNN embeddings, MC-PA-RWF features are fixed, explicit, and traceable to a physical channel, walk length, and within- or cross-group walk pattern. Thus, the proposed method offers competitive accuracy without relying on fully latent learned graph representations.

\subsubsection{Failure-Class (F1) Detection} 
Balanced accuracy summarises performance across both classes, but failure-class F1 provides a more application-relevant safety measure because the failure-prone class corresponds to operating scenarios that may lead to cascading outages. In this setting, missing a failure-prone graph is more consequential than misclassifying a stable one, so we separately evaluate how well each method identifies this critical class. Table~\ref{tab:failure_f1_gnn_sig} compares MC-PA-RWF+ against the strongest GNN baseline at the largest evaluated dataset size for each system. On IEEE24, failure-class F1 improves from $97.24\%$ for TransformerConv to $98.84\%$ for MC-PA-RWF+, a gain of $1.60$ percentage points ($p=0.0015$). On IEEE39, F1 improves from $91.03\%$ to $96.87\%$, a gain of $5.84$ percentage points ($p=0.0024$). On the UK benchmark, it improves from $96.95\%$ to $99.05\%$, a gain of $2.10$ percentage points ($p=0.0147$). All three improvements are statistically significant under paired two-sided $t$-tests at the $0.05$ level. The IEEE39 improvement is especially important. Although the strongest methods are close in balanced accuracy, MC-PA-RWF+ identifies failure-prone operating scenarios more effectively. This shows why F1 class-level metrics should be considered alongside balanced accuracy.



\section{Conclusion}
\label{sec:conclusion}

This work demonstrates that physically meaningful edge states provide essential
information for graph-level classification tasks for cyber-physical energy systems. Across the
IEEE24, IEEE39, and UK power system datasets (PowerGraph benchmarks), the proposed MC-PA-RWF variants
consistently improve over topology-only RWF and remain effective in settings
where node-feature RWF does not generalise well. At the largest evaluated
dataset sizes, MC-PA-RWF+ achieves balanced accuracies between $98.04\%$ and
$99.32\%$, matching or exceeding strong GNN baselines across the three systems.
More importantly, it improves failure-class F1 over the best GNN baseline by
$1.60$--$5.84$ percentage points, with statistically significant gains on every
benchmark. These results indicate that explicitly modelling edge-state-dependent
propagation improves not only overall classification performance but also the
detection of safety-critical failure-prone scenarios.

The channel-level analysis further shows that dynamic edge attributes reflecting changing operating conditions are more discriminative than static interaction properties alone. This confirms that MC-PA-RWF captures interpretable failure-related patterns from edge-level cyber-physical behaviour. 
MC-PA-RWF therefore offers a scalable and transparent alternative to fully
latent graph representations while preserving linear complexity with respect to the number of graph instances. Future work will investigate adaptive channel selection and extend
the framework to other cyber-physical domains with application-specific edge
information.

\section{Acknowldgement}
\label{}
This work is supported by the Deakin Mini ARC Analog Program. ChatGPT-5 is used to assist with language editing. All conceptual content and technical contributions are solely of the author.

\end{document}